\documentclass[letterpaper]{article} 
\usepackage{aaai2027}  
\usepackage[hyphens]{url}  
\usepackage{graphicx} 
\usepackage{natbib}  
\usepackage{caption} 
\usepackage[hyphens]{url}  
\usepackage{graphicx} 
\usepackage{natbib}  
\usepackage{caption} 
\usepackage{algorithm}
\usepackage{algorithmic}

\usepackage{amsmath,amssymb,amsfonts}
\usepackage{array}

\usepackage{url}
\usepackage{subcaption}
\usepackage{booktabs} %
\usepackage{multirow}
\usepackage{xcolor} %
\usepackage{algorithm}
\usepackage{enumitem}                            %

\newcommand{\code}[1]{\ifmmode\text{\ttfamily #1}\else\texttt{#1}\fi}
\usepackage{newfloat}
\usepackage{listings}
\DeclareCaptionStyle{ruled}{labelfont=normalfont,labelsep=colon,strut=off} 
\floatstyle{ruled}
\newfloat{listing}{tb}{lst}{}
\floatname{listing}{Listing}

\usepackage{booktabs}

\usepackage{microtype}
\usepackage{graphicx}
\usepackage{subcaption}
\usepackage{booktabs} 
\usepackage{multirow}
\usepackage{arydshln}
\usepackage{pifont}
\usepackage{booktabs}
\usepackage{subcaption} 

\usepackage{amsmath}
\usepackage{amssymb}
\usepackage{mathtools}
\usepackage{amsthm}
\usepackage{algorithm}
\usepackage{algorithmic}

\title{Task-State Adaptation with Prototype Memory for Multi-Task Dense Prediction}
\author{
    Yangyang Xu\textsuperscript{\rm 1},
    Haobo Yuan\textsuperscript{\rm 2},
    Yuzhu Wang\textsuperscript{\rm 1},
    Duo Su\textsuperscript{\rm 1},
    Xi Ye\textsuperscript{\rm 1},
    Yibo Yang\textsuperscript{\rm 3},
    Jun Zhu\textsuperscript{\rm 1}\thanks{Corresponding Author.}
}

\affiliations{
    \textsuperscript{\rm 1}Tsinghua University\\
    \textsuperscript{\rm 2}University of California, Merced\\
    \textsuperscript{\rm 3}Shanghai Jiao Tong University
}

\begin{document}

\maketitle


\begin{abstract}
Vision foundation backbones provide strong representations for dense prediction, yet a single shared feature still needs to support tasks with different, image-dependent adaptation requirements. We propose MemMTL, a multi-task dense prediction framework that estimates a compact task state from global visual context and refines it through a learnable task-state prototype memory. The refined state is converted into task-conditioned expert logits and combined with token-level logits before sparse top-$k$ selection over a local expert bank shared by all tasks. A separate task-agnostic residual bank provides a common adaptation path, and both paths are added once to the backbone feature before task-specific prediction. We specify a matched evaluation protocol on NYUD-v2 and PASCAL-Context with SAM~3 and ViT-L backbones to measure predictive quality, computational cost, and the contributions of task-state conditioning, prototype retrieval, and sparse routing. The numerical record in the present working draft predates this canonical implementation and must be regenerated before it can support empirical claims.
\end{abstract}

\section{Introduction}

Multi-task dense prediction learns semantic segmentation, depth estimation, surface normal prediction, boundary detection, and related pixel-level tasks in a unified model~\cite{MTL_survey_2021}. Recent vision backbones, including ViT~\cite{ViT2021} and SAM~3~\cite{sam3}, provide strong shared features for these tasks~\cite{InvPT_2022,yang2025multitask,mtl2025swiss}. A remaining design question is how to adapt the same dense feature to different tasks and images without replicating a large decoder for every output.

Existing methods approach this question from complementary directions. Optimization methods balance task losses or modify task gradients~\cite{2018MTL_uncertainty,chen2018gradnorm,yu2020_MTL_pcgrad,liu2021_MTL_CAgrad}. Architectural methods introduce task interaction modules~\cite{atrc_2021,InvPT_2022,xu2023demt,Huang2024mtlsem}, prompts~\cite{taskprompter2023}, adapters~\cite{jiang2024task}, or mixture-of-experts (MoE) layers~\cite{liang2022mvit,taskexpert23,yang2024multiMLoRE}. MoE designs are particularly attractive because they can separate reusable expert transformations from input-dependent routing. Their behavior, however, depends on how image-level task requirements and token-level evidence are combined when selecting experts.

We propose MemMTL, a task-state-conditioned sparse adaptation framework for multi-task dense prediction. MemMTL first pools the fused backbone feature and maps it to a coarse state for each image--task pair through a sparse mixture of global state experts. A learnable task-state prototype memory then retrieves a small set of normalized prototypes by cosine similarity and refines the coarse state through a residual connection and layer normalization. The refined state produces image-level expert logits. Rather than applying these logits after expert selection, MemMTL adds them to token-level logits before top-$k$ routing. Consequently, both global task context and local token evidence determine the experts that are dispatched.

All tasks use the same routed local expert bank, so an expert index has a consistent meaning across tasks. In parallel, a separate task-agnostic expert bank produces a shared residual from the dense feature. Both expert modules return residuals only, and the final task feature is the backbone feature plus one routed task residual and one task-agnostic residual. This construction makes the sharing pattern and residual path explicit while keeping the task losses and their balancing coefficients independent of the routing design.

We target matched evaluation with SAM~3~\cite{sam3} and ViT-L~\cite{ViT2021} on NYUD-v2~\cite{NYUD2012} and PASCAL-Context~\cite{pascal2014}. The required experiments measure the overall multi-task trade-off, routing sparsity, expert count, and the contributions of task-state conditioning and prototype retrieval. The numerical tables currently retained in the manuscript were generated before the canonical implementation and serve only as a rerun manifest, not as evidence for these claims.

Our main contributions are:
\begin{itemize}
    \item We formulate a hierarchical sparse router in which a prototype-refined, image-level task state and token-level visual evidence jointly determine top-$k$ expert selection.
    \item We introduce a learnable task-state prototype memory and an explicit two-residual design consisting of one task-routed expert bank shared across tasks and one task-agnostic expert bank.
    \item We define a matched, multi-seed evaluation protocol for two vision backbones on NYUD-v2 and PASCAL-Context, including component, routing, expert-count, efficiency, and intervention studies.
\end{itemize}

\section{Related Work}
\label{sec:relate}

\textbf{Multi-Task Dense Prediction under Strong Vision Backbones.}
Multi-task learning (MTL)~\cite{Pad-net_2018,MTL_survey_2021} for dense prediction jointly addresses pixel-level tasks such as semantic segmentation, depth, normals, and boundaries~\cite{MTL_survey_2021,InvPT_2022,xu2023demt,yang2025multitask}. 
A core challenge is negative transfer: semantic tasks prefer context-rich, category-discriminative features, while geometric tasks require spatial continuity and metric consistency. 
Existing approaches mitigate this via (i) optimization strategies ($e.g.,$ loss balancing and gradient correction~\cite{2018MTL_uncertainty,chen2018gradnorm,sener2018multi_MGDA,yu2020_MTL_pcgrad,liu2021_MTL_CAgrad,l2021_IMTL,li2026pike,mtlopt2026ntkmtl}) and (ii) architectural designs for task interaction and feature sharing~\cite{Pad-net_2018,Mti-net_2020,atrc_2021,InvPT_2022,taskprompter2023,xu2023multi,Huang2024mtlsem,yang2025multitask}. 
Recent work uses strong vision backbones with adapters and parameter-efficient modules~\cite{lu2024prompt,bhattacharjee2023vision,Polyhistor22,MTLoRA24,yang2024multiMLoRE,xu2023demt,xin2024vmt,jiang2024task,wang2025mtsam,mtlopt2026exploring}, showing that representation bottlenecks can be largely alleviated. In particular, ViT~\cite{ViT2021} and SAM~\cite{sam3} provide multi-scale features for dense understanding~\cite{kirillov2023sam,ravi2024sam2}. 
Our work focuses on a complementary regime: even with strong shared features, semantic and geometric tasks still require sample-specific adaptations. MemMTL therefore targets residual cross-task conflict beyond backbone representation limits.

\textbf{Conditional Routing for Task Conflict Modeling.}
Conditional computation balances sharing and specialization by activating different parameters for different inputs or tasks. MoE models~\cite{jacobs1991adaptive,jacobs1993learning,mu2025moesurvey,moe-tang2024} are a representative family and have been widely used in large-scale sparse modeling~\cite{fedus2022switch,du2022glam,dai2024deepseekmoe}. In MTL, MoE-style routing allocates computation across samples, tokens, or tasks~\cite{chen2023mod,liang2022mvit,taskexpert23,chen2023adamv,yang2024multiMLoRE}. For dense prediction~\cite{densePredic_fapn_2021}, such methods~\cite{taskexpert23,yang2024multiMLoRE,densePredic_cwkdis_2021} reduce interference compared with fully shared decoders. However, their routers are usually driven by task identity, local features, learned task memories, or task-specific controllers, without explicitly estimating the sample-level conflict state between semantic and geometric demands.
Beyond sparse routing, prompts, adapters, and task-conditioned controllers inject task cues into the backbone or decoder to specialize predictions for each output~\cite{taskprompter2023,jiang2024task,huang2024going,xu2023multi,lu2024prompt}.
Task identity alone cannot capture sample-dependent conflicts. MemMTL learns conflict-aware task states through prototype memory, while a shared path preserves task-invariant structure.

\section{Method}
\label{sec:method}

\begin{figure*}[t!]
\centering
  \includegraphics[width=0.93\textwidth]{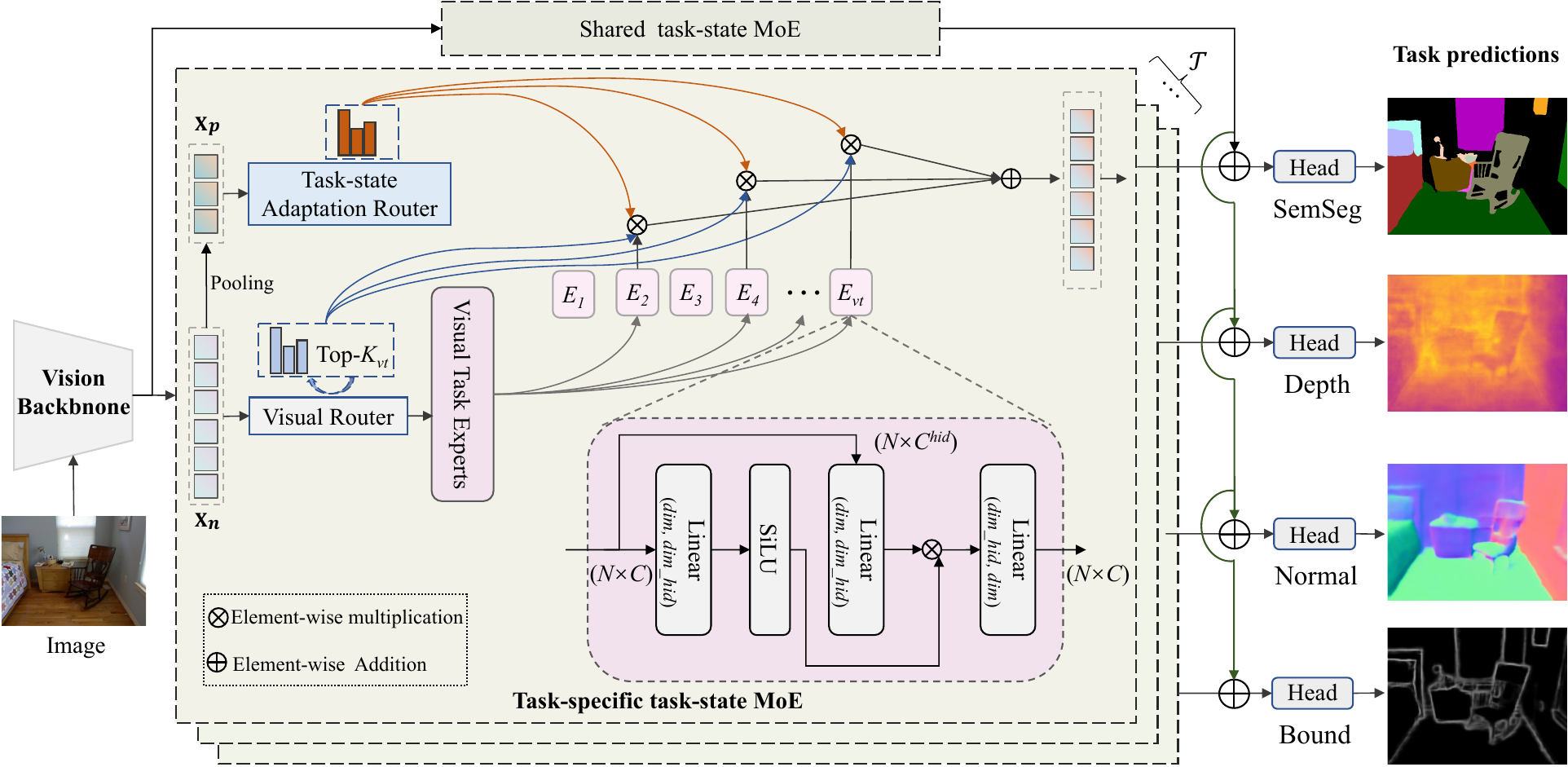}
 \caption{Framework of MemMTL. The backbone extracts multi-scale features shared across tasks. A task-specific task-state MoE models task-dependent residuals, while a shared task-state MoE preserves cross-task structure. 
}
\label{fig:over_view}
\end{figure*}

\subsection{Overview}

\textbf{Motivations}. Foundation vision backbones such as SAM~3 and ViT substantially alleviate the representation bottleneck in dense prediction, yet they do not remove the need for task-specific adaptation in multi-task settings. Semantic segmentation relies on category-discriminative and context-aware representations, whereas depth estimation, surface normal prediction, and boundary detection favor geometric consistency, local continuity, and fine structural cues. 
Even with a strong shared backbone, tasks may require conflicting sample-specific adaptations, resulting in \emph{residual cross-task task-state prototype}.

The overview of MemMTL is depicted in Figure~\ref{fig:over_view}. The framework is composed of three key components: the vision backbones encoder, task-specific task-state MoE module, and a shared task-state MoE module.
First, the vision backbone encoder ($i.e.,$ SAM 3 or ViT) processes the input image to produce generic multi-scale visual features that serve as a shared representation across tasks.
Second, the task-specific task-state MoE module introduces a learned task-state adaptation vector that dynamically gates local experts to adapt the shared features to heterogeneous dense prediction tasks.
Third, the shared task-state MoE module further refines these task-specific representations by modeling cross-task interactions and global spatial dependencies. The refined features are then passed through lightweight task-specific heads to produce the final dense predictions.

\subsection{Task-specific task-state MoE}

\textbf{Mixture-of-experts revisited.}
Mixture-of-experts (MoE) layers increase model capacity through conditional computation, where each input activates only a subset of expert modules~\cite{GShared21,2021scalingvmoe}. Given a set of experts $\{\mathcal{E}_i\}_{i=1}^{E}$, a router predicts an input-dependent gating distribution over experts. For an input feature $x$, a standard MoE layer can be written as
\begin{equation}\label{eq:moe_revisit}
y
=
\sum_{i=1}^{E}
g_i(x)\,\mathcal{E}_i(x),
\end{equation}
where $g_i(x)$ denotes the routing weight assigned to the $i$-th expert. In sparse MoE layers, $g_i(x)$ is typically obtained by applying a softmax over router logits and retaining only the top-$k$ experts, enabling efficient conditional specialization.

\textbf{Task-state MoE.}
For multi-task dense prediction, token-level routing alone is insufficient: it selects experts according to local feature patterns, but does not explicitly capture the global task demand induced by the current image. We therefore introduce a task-state adaptation MoE module on top of the fused foundation features from SAM~3 or ViT. The module combines two complementary routing signals: a visual token router and a task-state adaptation router. The visual router produces token-wise sparse weights and top-$k$ expert indices from local features, following the common design of MTL MoE methods such as TaskExpert~\cite{taskexpert23}. In parallel, the task-state adaptation router predicts task-specific expert coefficients from the memory-refined task-state.

Let $\mathbf{x}_n$ denote the feature of spatial token $n$ as input to the task-specific task-state MoE module. The visual router outputs token-level weights $\omega_{n,i}$, while the task-state adaptation router outputs a task-level expert gate ${g}_t$. The task-specific task-state MoE update is
\begin{equation}\label{eq:task_state_moe}
\Delta^{\mathrm{ts}}_{t,n} = \sum_{i=1}^{E} \omega_{n,i}\, g_{t,i}\, \mathcal{E}_i(\mathbf{x}_n),
\end{equation}
where $\omega_{n,i}=0$ for experts not selected by the token-level top-$k$ router. The final task-specific feature is obtained by a residual update:
\begin{equation}
\mathbf{z}_{t,n} = \mathbf{x}_n + \Delta^{\mathrm{ts}}_{t,n}.
\end{equation}

This design separates two routing roles. The visual router determines \emph{where} specialized computation is needed at the token level, while the task-state adaptation router determines \emph{which} experts should be emphasized for the current task and image. As a result, our MoE supports fine-grained spatial specialization while maintaining global task consistency.

\subsection{Task-state adaptation router}

\begin{figure*}[!t]
\centering
  \includegraphics[width=0.930\textwidth]{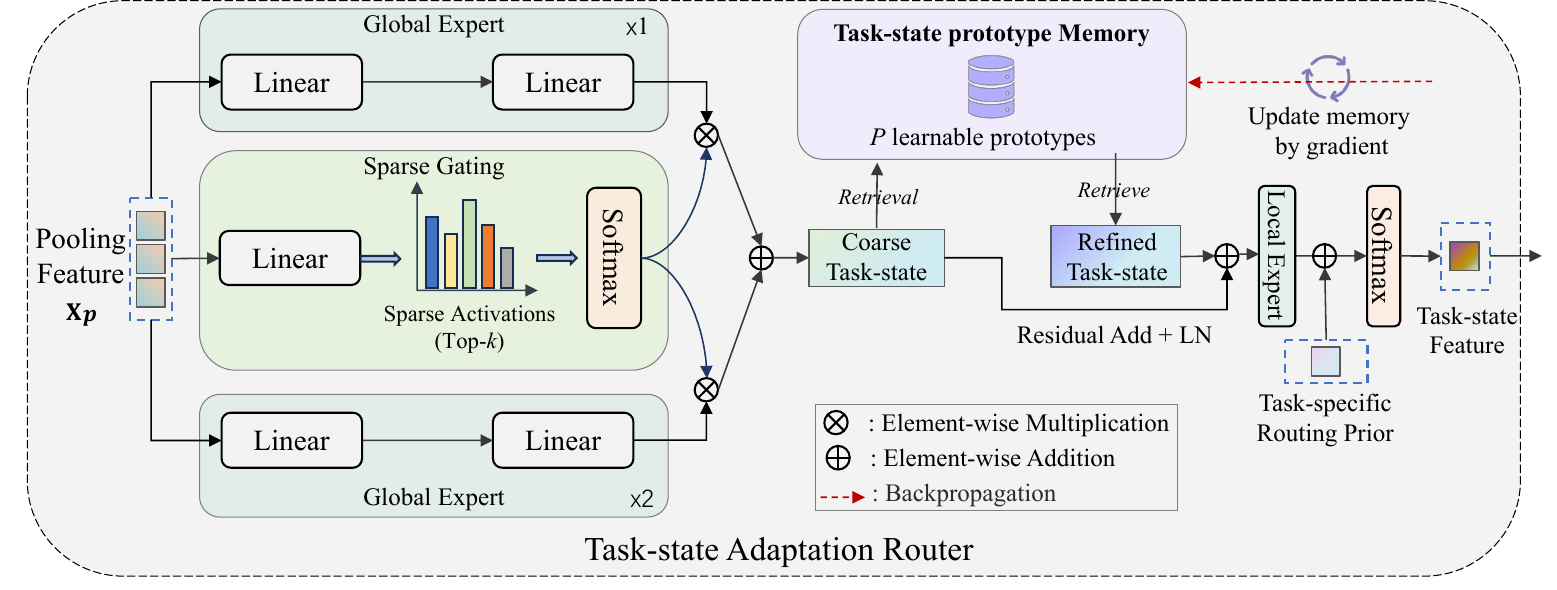}
  \caption{Task-state adaptation router. Pooling feature is mapped to a sample-conditioned coarse task-state through a sparse mixture of global experts, refined by a task-state prototype memory, and then used to modulate downstream local routing.}\label{fig:language_moe}
\end{figure*}

To effectively integrate task semantics into the expert routing process, we introduce a task-state adaptation router, as shown in Figure~\ref{fig:language_moe}.
A gate computes token-wise weights and top-\textit{k} expert indices based on visual features, following recent MTL MoE method (TaskExpert~\cite{taskexpert23}). Simultaneously, the task-state adaptation router provides task-aware coefficients for each expert.
During expert computation, outputs are modulated by both token-level features and task-state features. This synergy allows fine-grained specialization (per-token routing) while maintaining global task consistency.

\textbf{Coarse task-state from global experts.}

$\mathbf{x}_p$ is obtained by applying a pooling operation to $\mathbf{x}_n$.
Let $\mathcal{T}$ be the set of tasks. For each task $t \in \mathcal{T}$, the pooling feature $\mathbf{x}_p$ are fed to global expert logits:
\begin{equation}
\mathbf{r}_t({\mathbf{x}_p}) = \mathbf{W}^{(r)}_{t}{\mathbf{x}_p} \in \mathbb{R}^{E_g},
\end{equation}
where $E_g$ is the number of global experts and $\mathbf{W}^{(r)}_{t}$ is a task-specific linear projection. We then perform top-$k$ sparse normalization to obtain global mixture weights:
\begin{equation}
\boldsymbol{\pi}_t = \mathrm{TopKSoftmax}\!\left(\mathbf{r}_t({\mathbf{x}_p}), {ek}_g\right)\in\mathbb{R}^{E_g},
\label{eq:global_gate}
\end{equation}
where ${ek}_g$ is the number of activated global experts.

The global experts $\{G_i\}_{i=1}^{E_g}$ are small MLPs operating on the pooling feature:
\begin{equation}
G_i(\mathbf{p}) = \mathbf{W}^{(2)}_{i}\,\sigma\!\left(\mathbf{W}^{(1)}_{i}{\mathbf{x}_p}\right)\in\mathbb{R}^{D_s},
\end{equation}
where $\sigma(\cdot)$ is GELU and $D_s$ is the coarse task-state dimension.
The coarse task-state vector is obtained by a gated mixture of global experts:
\begin{equation}
\tilde{\mathbf{s}}_t= \sum_{i=1}^{E_g} \pi_{t,i}\, G_i({\mathbf{x}_p}) \in \mathbb{R}^{D_s}.
\label{eq:coarse_task_state}
\end{equation}
This state summarizes the image-level visual evidence relevant to task $t$ before memory refinement.

\textbf{Task-state prototype memory.}
Dense prediction datasets are relatively small, and many images exhibit recurring task-state prototype patterns such as semantic ambiguity near object boundaries or geometry-sensitive indoor layouts~\cite{MTL_survey_2021}. MemMTL represents these recurring patterns with a learnable task-state prototype memory:
\begin{equation}
\mathbf{M}=[\mathbf{m}_1;\ldots;\mathbf{m}_P]\in\mathbb{R}^{P\times D_s},
\end{equation}
where each prototype $\mathbf{m}_j$ represents a latent task-state prototype mode. 
Unlike a queue or dataset-level cache, $\mathbf{M}$ is a small set of trainable prototypes learned end-to-end with the multi-task objective.
For each task, a task-specific query projection maps the coarse task-state to a memory query:
$\mathbf{q}_t=\mathbf{W}^{(q)}_t\tilde{\mathbf{s}}_t\in\mathbb{R}^{D_s}.$
We normalize the query and prototype rows,
$\hat{\mathbf{q}}_{t}=\mathbf{q}_{t}/
\max(\|\mathbf{q}_{t}\|_2,\epsilon)$ and
$\hat{\mathbf{m}}_j=\mathbf{m}_j/
\max(\|\mathbf{m}_j\|_2,\epsilon)$.
Prototype retrieval is performed by a temperature-scaled sparse softmax:
\begin{equation}
\boldsymbol{\rho}_t
=\mathrm{TopKSoftmax}\!\left(\frac{\mathbf{\hat{q}}_t\mathbf{\hat{M}}^{\top}}{\tau}, k_p\right)
\in\mathbb{R}^{P},
\label{eq:prototype_retrieval}
\end{equation}
where $\hat{\mathbf{M}}$ stacks the normalized prototypes. $P$ is the number of memory prototypes, $\tau$ is the temperature, and $k_p$ is the number of retrieved prototypes. The retrieved refined task-state is
\begin{equation}
\mathbf{c}_{t}
=
\hat{\mathbf{M}}^\top\boldsymbol{\rho}_{t}
\in\mathbb{R}^{D_s}.
\label{eq:prototype_context}
\end{equation}
The refined task-state used for routing is a normalized residual:
\begin{equation}
\mathbf{s}_t=\mathrm{LN}\!\left(\tilde{\mathbf{s}}_t+\gamma\mathbf{c}_t\right),
\label{eq:memory_refined_state}
\end{equation}
where $\gamma$ controls the contribution of retrieved context.
Since $\mathbf{M}$ is an ordinary model parameter, it is updated by the same optimizer as the router and experts.

\textbf{Task-specific routing prior.}
Given the memory-refined task-state $\mathbf{s}_t$, we predict logits over $E$ local experts via a task-specific linear map:
$\mathbf{u}_t = \mathbf{W}^{(u)}_{t}\mathbf{s}_t \in\mathbb{R}^{E}.
\label{eq:local_logits}$
Sparse expert routing can be unstable early in training on small dense benchmarks, so we add a weak task-specific prior $\boldsymbol{\alpha}_t\in\mathbb{R}^{E}$:
\begin{equation}
\tilde{\mathbf{u}}_t = \mathbf{u}_t + \lambda\,\boldsymbol{\alpha}_t,
\label{eq:bootstrap_prior}
\end{equation}
where $\lambda$ can be annealed during training.
Here, $\boldsymbol{\alpha}_t$ introduces a weak semantic–geometry bias while preserving sample adaptivity. The resulting task-conditioned local expert gate is
\begin{equation}
\mathbf{g}_t = \mathrm{Softmax}(\tilde{\mathbf{u}}_t)\in\mathbb{R}^{E}.
\label{eq:local_gate}
\end{equation}
For each task $t$, the router outputs $(\mathbf{s}_t,\mathbf{g}_t,\boldsymbol{\rho}_t)$: the refined task-state, the local expert preference, and the retrieved task-state prototype distribution.

\subsection{Shared task-specific MoE}

\textbf{Shared task-specific MoE branch.}
MemMTL uses a shared task-specific MoE to preserve structure that remains useful across tasks. This branch intentionally uses a uniform task-level gate, so it is not another task-conditioned controller:
\begin{equation}\label{eq:shared_moe_path}
\Delta^{\mathrm{sh}}_{n}
=
\sum_{i=1}^{E}
\omega^{\mathrm{sh}}_{n,i}\,\bar{g}_{i}\, \mathcal{E}^{\mathrm{sh}}_i(\mathbf{x}_n),
\qquad
\bar{g}_{i}=\frac{1}{E}.
\end{equation}
Note that the shared branch is not a non-routing branch: token-level weights
$\omega^{\mathrm{sh}}_{n,i}$ still route spatial tokens to experts. The uniform
$\bar{g}_i$ only removes task-level conditioning, keeping this branch
task-state prototype.
The final task-adapted feature at token $n$ is
\begin{equation}\label{eq:ourmoe8}
\mathbf{z}_{t,n}
=
\mathbf{x}_n
+
\Delta^{\mathrm{sh}}_{n}
+
\Delta^{\mathrm{ts}}_{t,n}.
\end{equation}
The residual connection stabilizes routing by preserving the original features. Reshaping $\{\mathbf{z}_{t,n}\}$ into $\mathbf{F}_t$, the task head predicts $\hat{\mathbf{y}}_t=H_t(\mathbf{F}_t)$.

\subsection{Training Objective}

Training MemMTL optimizes a standard multi-task objective over all dense prediction heads:
\begin{equation}\label{equ:loss}
 \begin{aligned}
    {\mathcal L_{mt}} &= \sum_{t\in\mathcal{T}} \beta_{t}{\mathcal L_{t}},
       \end{aligned}
\end{equation}
where $\beta_t$ denotes the task-balancing coefficient and $\mathcal{L}_{t}$ is the loss of task $t$.

\section{Experiments}
\label{sec:exp}

\begin{table*}[!t]
\centering
\footnotesize
\caption{Results on PASCAL-Context. ST baseline denotes single task baseline. `$\downarrow$': lower is better. `$\uparrow$': higher is better. $\Delta_m$ denotes the average per-task performance drop (the higher, the better).}
\label{tab:sota_comparison_pascal}
\setlength{\tabcolsep}{4.99pt}
\begin{tabular}{lcccccccll}
\toprule[0.1em]
 \multirow{2}*{Model}  &\multirow{2}*{Backbone}  &{SemSeg}   &PartSeg &Sal  &Normal &Bound &\multirow{2}*{$\Delta_m[\%]$$\uparrow$}  &FLOPs &Params\\
     &  & (mIoU)$\uparrow$  & (mIoU)$\uparrow$  &(maxF)$\uparrow$  &(mErr)$\downarrow$  &(odsF)$\uparrow$ & &(G)$\downarrow$ &(M)$\downarrow$\\
\hline
ST baseline  &HRNet18 &62.23 &61.66 &85.08 &13.69 &73.06 &0.00 &- &- \\
MTI-Net &HRNet18 &61.70 &60.18 &84.78 &14.23 &70.80 &-2.10  &161 &128\\
ATRC    &HRNet18 &57.89 &57.33 &83.77 &13.99 &69.74 &-4.45 &216 &96\\
DeMT    &HRNet18 &59.23 &57.93 &83.93 &14.02 &69.80 &-3.79  &- &-\\
\hdashline
ST baseline            &ViT-L &81.62 &72.21 &84.34 &13.59 &76.79 &0.00    &- &317 \\
PAD-Net        &ViT-L &78.01 &67.12 &79.21 &14.37 &72.60 &-5.72 &773  &330\\
MTI-Net        &ViT-L &78.31 &67.40 &84.75 &14.67 &73.00 &-4.62 &774  &851\\
ATRC~          &ViT-L &77.11 &66.84 &81.20 &14.23 &72.10 &-5.50 &871  &340\\
InvPT~         &ViT-L &79.03 &67.61 &84.81 &14.15 &73.00 &-3.61 &669  &423\\
TaskPrompter~  &ViT-L &80.89 &68.89 &84.83 &13.72 &73.50 &-2.03 &497 &401\\
TaskExpert~    &ViT-L &80.64 &69.42 &84.87 &13.56 &73.30 &-1.74 &622 &420\\
MLoRE          &ViT-L &81.41 &70.52 &84.90 &13.51 &75.42 &-0.64 &571 &407\\
MemMTL (Ours)   &ViT-L &80.12 &69.81 &84.30 &13.63 &75.40 &-1.40 &451 &327\\
\hdashline
ST baseline            &SAM 3 &81.48 &75.62 &83.72 &13.71 &76.90 &0.00	&- &454 \\
MemMTL (Ours)  &SAM 3 &80.59 &74.68 &83.21 &13.91 &76.70 &-0.94 &612 &461 \\
\bottomrule[0.1em]
  \end{tabular}
\end{table*}

\textbf{Datasets.}
We evaluate MemMTL on NYUD-v2~\cite{NYUD2012} and PASCAL-Context~\cite{pascal2014}. 
NYUD-v2 contains 1,449 RGB-D images (795 train / 654 test) from indoor scenes, with annotations for semantic segmentation (SemSeg), depth estimation (Depth), surface normal prediction (Normal), and boundary detection (Bound). 
PASCAL-Context includes 10,103 images (4,998 train / 5,105 val), covering SemSeg,  human part segmentation (PartSeg), saliency detection (Sal), Normal, and Bound.

\textbf{Evaluation metric.}
We evaluate with task-specific metrics: mIoU for SemSeg and PartSeg, RMSE for Depth, mErr for Normal, odsF for Bound, and maxF for Sal. 
We further report the average relative performance drop $\Delta_m$ to summarize multi-task trade-offs:
$\Delta_m=\frac{1}{\mathcal{T}} \sum_{t=1}^{\mathcal{T}} (-1)^{l_t}\frac{F_{m,t}-F_{s,t}}{F_{s,t}} \times 100\%$,
where $F_{m,t}$ and $F_{s,t}$ denote multi-task and single-task performance on task $t$.
{$l_t$=1/0} denotes lower/higher is better.

\textbf{Implementation Details.}
We implement MemMTL in PyTorch using a SAM~3 backbone with feature-pyramid outputs. The multi-scale features are upsampled, concatenated, and projected by a $1\times1$ conv-BN-GELU block into a 352-channel feature map. We also evaluate MemMTL with a ViT-L backbone to test generality.
The task-state prototype memory uses 8 ($i.e., P$=8) learnable prototypes with top-2 retrieval, whose retrieved state is added to the coarse task-state before normalization. For MemMTL-CPA, the loss weights for usage, diversity, capacity alignment, and router-expert coupling are $0.01$, $0.001$, $0.01$, and $0.005$. Training uses bs 2, learning rate $2\times10^{-5}$, 6 local experts, and Adam. Experiments are conducted on NVIDIA A800-SXM4-80GB GPUs.

\subsection{Main Results}

\textbf{Performance on PASCAL-Context.}
Table~\ref{tab:sota_comparison_pascal} reports results on the PASCAL-Context dataset across five dense prediction tasks under different backbones. 
When paired with a ViT-L encoder, MemMTL provides a favorable efficiency-accuracy trade-off: compared with MLoRE, it uses fewer parameters and lower FLOPs while maintaining competitive performance across all five dense prediction tasks. This indicates that the proposed routing design can reduce the cost of multi-task adaptation without relying on a heavier task-specific parameterization (See Appendix Table~S2 for the parameter decomposition).
More notably, with the stronger SAM~3 backbone, MemMTL further narrows the gap to the single-task baseline. Although ST baseline trains one independent model per task and therefore serves as a strong upper-bound reference, MemMTL achieves only a $0.94\%$ average performance drop while using a unified multi-task model. Notably, MemMTL preserves strong performance on structure-sensitive tasks such as part segmentation and boundary detection, reaching $74.68$ mIoU on PartSeg and $76.70$ on Bound.
These results suggest that prototype-refined task routing enables efficient cross-task sharing while preserving task specialization.

\begin{table}[t!]
\footnotesize
\caption{Results on NYUD-v2. $\uparrow$/$\downarrow$: higher/lower is better. \label{tab:stoa_nyud_moe}}
\setlength{\tabcolsep}{2pt}
\centering
\begin{tabular}{lccccccc}
\toprule[0.1em]
 \multirow{2}*{Model}  &backbone &SemSeg &Depth & Normal & Bound &{$\Delta_m$}\\
       &(mIoU)$\uparrow$   &(rmse)$\downarrow$   & (mErr)$\downarrow$  &(odsF)$\uparrow$ &[\%]\\
\hline
ST baseline            &ViT-L   &56.77 &0.5141 &18.56 &78.93 &0.00\\
InvPT          &ViT-L   &53.56 &0.5183 &19.04 &78.10 &-2.52\\
TaskPrompter   &ViT-L   &55.30 &0.5152 &18.47 &78.20 &-0.81\\
TaskExpert     &ViT-L   &55.35 &0.5157 &18.54 &78.40 &-0.84\\
MLoRE          &ViT-L   &55.96 &0.5076 &18.33 &78.43 &0.11\\
MemMTL (Ours)   &ViT-L   &56.81 &0.4981 &17.97 &79.00 &1.52\\
\hdashline
ST baseline            &SAM 3   &61.26 &0.4531 &16.31 &80.10 &0.00\\
MemMTL (Ours)   &SAM 3  &60.28 &0.4655 &17.03 &79.80 &-2.15\\
\bottomrule[0.1em]
  \end{tabular}
\end{table}

\textbf{Performance on NYUD-v2.}
Table~\ref{tab:stoa_nyud_moe} reports the comparison on NYUD-v2, which contains both semantic and geometry-oriented dense prediction tasks. With a ViT-L backbone, MemMTL consistently improves over prior multi-task methods and even surpasses the single-task baseline on average, achieving a positive $\Delta_m$ of $1.52\%$. 
These gains suggest that the proposed task-state prototype routing is effective for balancing heterogeneous task demands in indoor scenes. When equipped with the stronger SAM~3, MemMTL remains close to the single-task upper-bound while using one unified multi-task model. Compared with the SAM~3 ST baseline, the remaining gap is mainly caused by geometry-sensitive tasks such as depth and surface normal estimation, where single-task optimization can allocate the full model capacity to a single prediction target. 
This highlights a key challenge in SAM~3 MTL: stronger representations narrow the gap, but explicit cross-task task-state prototype resolution remains necessary.

\subsection{Ablation Study}

\textbf{Ablation on components.}
Table~\ref{tab:ablation_component} ablates the main components of MemMTL on NYUD-v2. Starting from the multi-branch baseline, adding the task-specific task-state MoE (TTM) improves semantic segmentation, depth estimation, and boundary detection, showing the benefit of task-conditioned expert specialization. Adding the shared task-state MoE (STM) brings further gains, especially for geometry-related tasks such as depth and surface normals, suggesting that shared experts capture transferable structure across tasks. Removing the task-state adaptation router (w/o TAR) causes a clear drop from the model, indicating that expert capacity alone is insufficient without task-state-dependent routing. Overall, MemMTL achieves the best results across all metrics, improving the baseline by $4.90$ mIoU on semantic segmentation, reducing depth by $0.0358$, reducing normal error by $1.08$, and improving boundary by $2.6$. This confirms that TTM, STM, and TAR provide complementary gains.

\begin{table}[!t]
  \centering
  \footnotesize
  \setlength{\tabcolsep}{3.5pt}
  \caption{Ablation of the modules on NYUD-v2 with SAM~3. TTM denotes task-specific task-state MoE, STM is shared task-state MoE, and TAR is task-state adaptation router.}
  \label{tab:ablation_component}
  \begin{tabular}{lcccc}
    \toprule[0.1em]
    \multirow{2}{*}{Setting}
    & SemSeg
    & Depth
    & Normal
    & Bound \\
    & (mIoU)$\uparrow$
    & (RMSE)$\downarrow$
    & (mErr)$\downarrow$
    & (odsF)$\uparrow$ \\
    \noalign{\smallskip}
    \hline
    \noalign{\smallskip}
    MT baseline        & 55.38 & 0.5013 & 18.11 & 77.2 \\
    \textit{w/} TTM    & 57.05 & 0.4916 & 18.21 & 78.1 \\
    \textit{w/} STM    & 57.42 & 0.4868 & 17.74 & 78.6 \\
    \textit{w/o} TAR   & 58.96 & 0.4742 & 17.38 & 79.1 \\
    MemMTL              & 60.28 & 0.4655 & 17.03 & 79.8 \\
    \bottomrule[0.1em]
  \end{tabular}
\end{table}

\begin{figure*}[!t]
  \centering
  \begin{subfigure}[t]{0.27\textwidth}
    \centering
    \includegraphics[page=1,width=\linewidth]{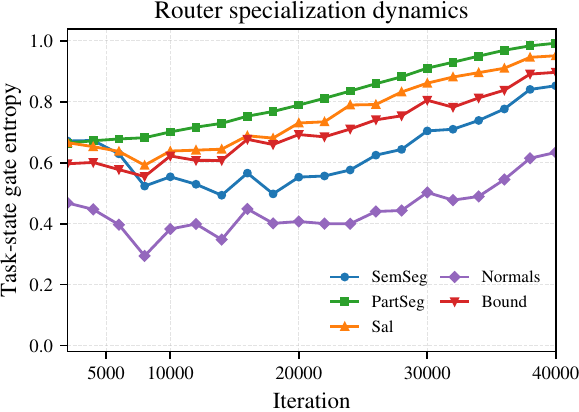}
    \caption{Routing dynamics}
  \end{subfigure}\hfill
  \begin{subfigure}[t]{0.27\textwidth}
    \centering
    \includegraphics[page=1,width=\linewidth]{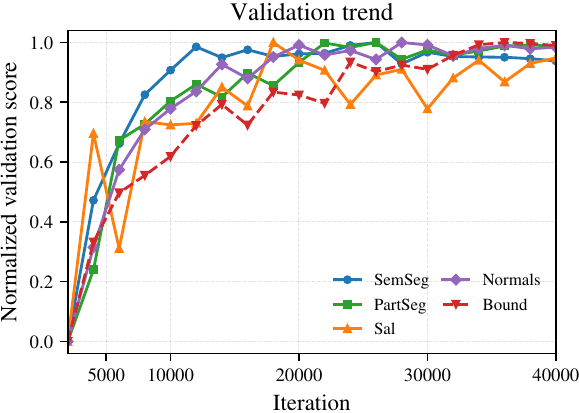}
    \caption{Validation trend}
  \end{subfigure}\hfill
  \begin{subfigure}[t]{0.45\textwidth}
    \centering
    \includegraphics[page=1,width=\linewidth]{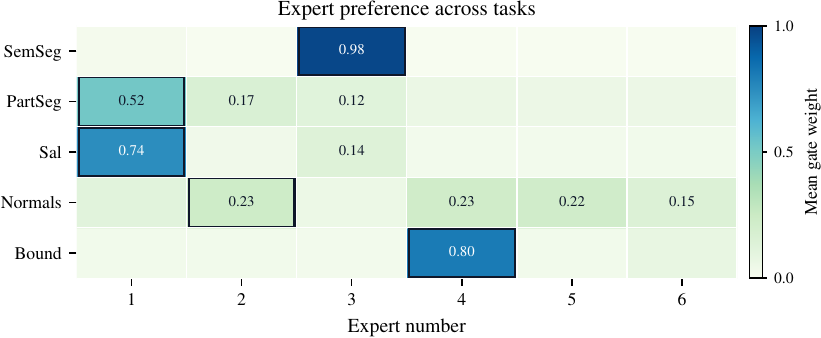}
    \caption{Task-specific expert preference}
  \end{subfigure}
  \caption{ (a) Routing dynamics. Task-state gate entropy reveals task-specific expert allocation patterns during training. (b) Validation trend. Performance improves with routing evolution, suggesting coupled expert allocation and multi-task optimization. (c) Task-specific expert preference. Mean gate weights show distinct expert preferences across tasks, with some tasks selecting dominant experts and others using more distributed mixtures.}
  \label{fig:routingdist}
\end{figure*}

\begin{table}[!t]
  \centering
  \footnotesize
  \setlength{\tabcolsep}{4.5pt}
  \caption{Ablation study of Top-\textit{k} in the task-state adaptation router on NYUD-v2 using the SAM~3 backbone. Top-\textit{k} denotes the number of sparse activations in the global router.}
  \label{tab:ab:task_state_router}
  \begin{tabular}{ccccc}
    \toprule[0.1em]
    \multirow{2}{*}{Top-\textit{k}}
    & SemSeg
    & Depth
    & Normal
    & Bound \\
    & (mIoU)$\uparrow$
    & (RMSE)$\downarrow$
    & (mErr)$\downarrow$
    & (odsF)$\uparrow$ \\
    \noalign{\smallskip}
    \hline
    \noalign{\smallskip}
    1 & 59.34 & 0.4786 & 17.42 & 79.1 \\
    2 & 60.28 & 0.4655 & 17.03 & 79.8 \\
    3 & 59.86 & 0.4708 & 17.18 & 79.5 \\
    4 & 59.57 & 0.4749 & 17.31 & 79.3 \\
    \bottomrule[0.1em]
  \end{tabular}
\end{table}

\textbf{Effect of Top-$k$ ($k_g$) in the task-state guided router.}
Table~\ref{tab:ab:task_state_router} studies the number of sparse activations in the global task-state router. Using only one activated state ($k=1$) leads to inferior results, suggesting that overly sparse routing restricts the model from capturing complementary task contexts. Increasing the value to $k=2$ achieves the best overall performance, with the highest semantic segmentation and boundary scores as well as the lowest depth and normal errors. This indicates that selecting a small set of task-state experts provides sufficient flexibility while preserving routing selectivity. Further increasing the number of activated states to $k=3$ or $k=4$ slightly degrades performance. 
Activating too many states makes routing less selective, mixing heterogeneous task contexts and weakening expert specialization. Moderate sparsity is therefore important for MemMTL, with $k=2$ best balancing task-specific adaptation and cross-task sharing.

\textbf{Effect of the number of experts.}
Table~\ref{tab:ablation_expernum} studies the number of experts in the task-specific task-state MoE. Using only four experts gives weaker results, suggesting insufficient capacity for modeling heterogeneous semantic, geometric, and boundary cues. Increasing the number to six improves performance across tasks, while eight experts further benefits semantic segmentation and depth estimation. However, using sixteen experts degrades depth and surface normal estimation, likely because routing becomes over-fragmented and each expert receives fewer effective updates. 
Overall, $6$--$8$ experts offer the best capacity--specialization trade-off.

\begin{table}[!t]
  \centering
  \footnotesize
  \setlength{\tabcolsep}{3.5pt}
  \caption{Ablation on the number of experts in the task-specific task-state MoE on NYUD-v2 using the SAM~3 backbone. The number of experts is set to 6 in all other experiments.}
  \label{tab:ablation_expernum}
  \begin{tabular}{ccccc}
    \toprule[0.1em]
    \multirow{2}{*}{Expert number}
    & SemSeg & Depth & Normal & Bound \\
    & (mIoU)$\uparrow$ & (RMSE)$\downarrow$ & (mErr)$\downarrow$
    & (odsF)$\uparrow$ \\
    \hline
    4  & 58.95 & 0.4725 & 17.29 & 79.8 \\
    6  & 60.28 & 0.4655 & 17.03 & 79.8 \\
    8  & 60.92 & 0.4627 & 17.04 & 79.8 \\
    16 & 59.46 & 0.4715 & 17.32 & 79.4 \\
    \bottomrule[0.1em]
  \end{tabular}
\end{table}

\begin{table}[!t]
  \centering
  \footnotesize
  \setlength{\tabcolsep}{3pt}
  \caption{Ablation of task-state prototype memory (TPM), including task-state-aware update and retrieval for task-state routing.}
  \label{tab:memory_intervention}
  \begin{tabular}{lcccc}
    \toprule[0.1em]
    \multirow{2}{*}{Setting}
    & SemSeg & Depth & Normal & Bound \\ & (mIoU)$\uparrow$ & (RMSE)$\downarrow$ & (mErr)$\downarrow$ & (odsF)$\uparrow$ \\
    \hline
    MT baseline                 & 55.38 & 0.5013 & 18.11 & 77.2 \\
    \textit{w/} task-state prototype update
                                & 58.72 & 0.4791 & 17.46 & 78.9 \\
    \textit{w/} task-state prototype retrieval
                                & 59.34 & 0.4746 & 17.25 & 79.2 \\
    MemMTL \textit{w/} TPM        & 60.28 & 0.4655 & 17.03 & 79.8 \\
    \bottomrule[0.1em]
  \end{tabular}
\end{table}

\textbf{Effect of the task-state prototype memory. }
Table~\ref{tab:memory_intervention} ablates the proposed task-state prototype memory. Adding task-state prototype updates in task-state prototype memory already improves all tasks over the MT baseline, indicating that recording recurring task-state prototype patterns provides useful task-state regularization. Enabling task-state prototype retrievals further improves performance by using the retrieved prototypes to guide routing. The full model achieves the best results, showing that memory update and retrieval are complementary for task-state adaptation.
We provide the detailed implementation and analysis of the task-state prototype memory in the Appendix~2.3.

\subsection{Task-State adaptation Routing Analysis}
We analyze the task-state adaptation router on PASCAL-Context in Figure~\ref{fig:routingdist}. 
Figure~\ref{fig:routingdist}(a) shows the entropy of task-state gates over local experts, where lower entropy indicates more concentrated routing. The tasks follow distinct entropy patterns, suggesting that different dense prediction objectives require different levels of expert specialization.
Figure~\ref{fig:routingdist} (b) shows that validation performance improves as routing evolves, indicating that expert allocation is learned jointly with multi-task optimization. 
Figure~\ref{fig:routingdist}(c) visualizes the mean gate weights at the selected checkpoint. 
SemSeg, Sal, and Bound favor distinct dominant experts, whereas Normal uses a more distributed mixture. This indicates that the task-state router learns structured, task-dependent expert allocation.

\subsection{Qualitative Analysis}
\label{sec:qualitative_analysis}

\begin{figure}[!t]
\centering
  \includegraphics[width=0.494\textwidth]{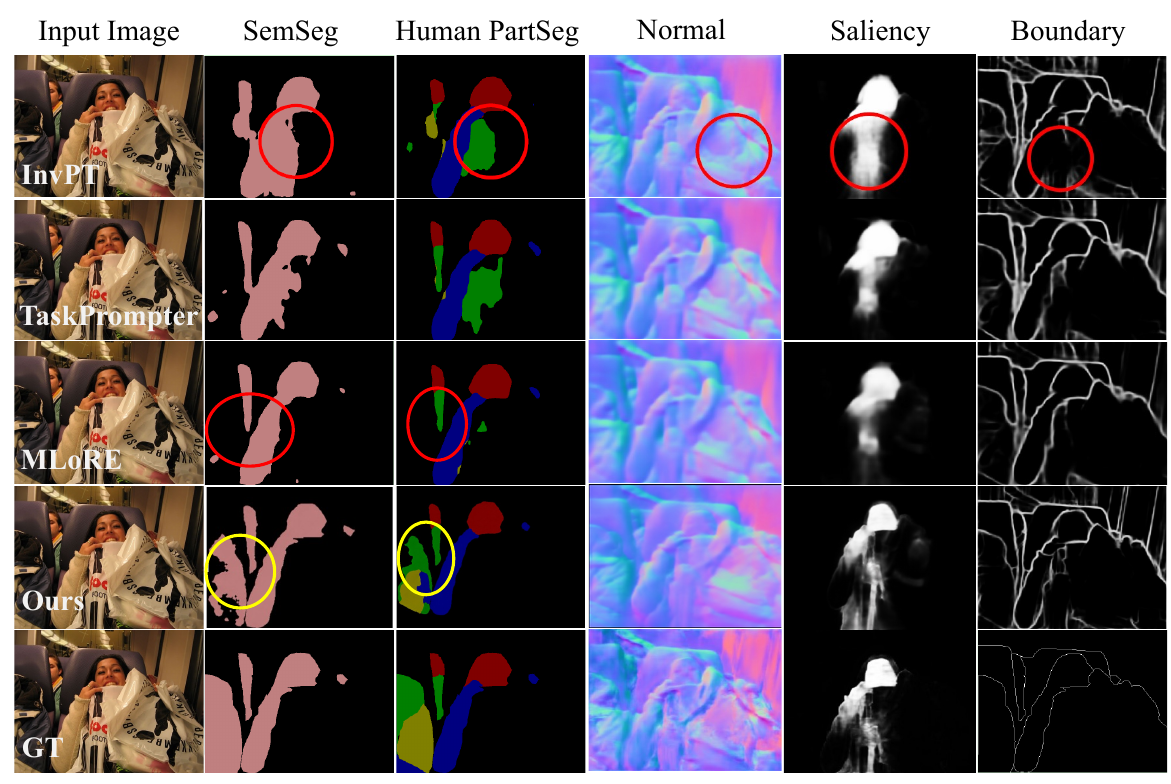}
\caption{Qualitative comparison on the five-task PASCAL-Context benchmark with InvPT~\cite{InvPT_2022}, TaskPrompter~\cite{taskprompter2023}, and MLoRE~\cite{yang2024multiMLoRE}. The columns show the input image and predictions for semantic segmentation, human-part segmentation, surface normals, saliency, and boundaries. Red circles indicate representative errors in competing methods, whereas yellow circles highlight regions where our predictions more closely match the ground truth, particularly in preserving semantic separation and fine-grained human-part structures. Best viewed in color and under magnification.}\label{fig:Qualitative}
\end{figure}

Figure~\ref{fig:Qualitative} presents a challenging scene with substantial foreground occlusion. For semantic segmentation, our method better preserves the separation and spatial extent of adjacent regions, whereas competing methods merge or omit parts near the person--seat interface. For human-part segmentation, it recovers finer structures and clearer part transitions, while InvPT and MLoRE exhibit missing or merged components in the highlighted areas. Our predictions also retain more localized surface-normal variations, more complete saliency regions, and more continuous boundaries. The consistent object extent across tasks suggests improved cross-task coherence. These examples complement, rather than replace, the quantitative evaluation.

\section{Conclusion}
\label{sec:conclusion}

We presented MemMTL, a task-state-conditioned sparse adaptation framework for multi-task dense prediction. MemMTL derives an image-conditioned state for each task and refines it by retrieving normalized vectors from a learnable task-state prototype bank. The refined state is combined with token-level evidence before sparse top-$k$ routing, allowing global task requirements and local visual content to jointly determine expert dispatch. A single routed expert bank is shared across tasks, while an independent task-agnostic bank provides a common residual; both residual paths are combined with the backbone feature through a single identity addition. This modular design is compatible with standard task objectives and does not depend on a particular loss-balancing strategy. Future work will investigate robustness to prototype and routing configurations, as well as transfer to unseen tasks and datasets. Experiments on NYUD-v2 and PASCAL-Context show that MemMTL achieves competitive multi-task performance with a favorable accuracy--efficiency trade-off while producing spatially coherent predictions across heterogeneous dense prediction tasks.

\bibliography{aaai2027}


\end{document}